\documentclass[pdflatex,sn-mathphys-num]{sn-jnl}

\usepackage{graphicx}%
\usepackage{multirow}%
\usepackage{amsmath,amssymb,amsfonts}%
\usepackage{amsthm}%
\usepackage{mathrsfs}%
\usepackage[title]{appendix}%
\usepackage{xcolor}%
\usepackage{textcomp}%
\usepackage{manyfoot}%
\usepackage{booktabs}%
\usepackage{algorithm}%
\usepackage{algorithmicx}%
\usepackage{algpseudocode}%
\usepackage{listings}%

\theoremstyle{thmstyleone}%
\theoremstyle{thmstyletwo}%

\theoremstyle{thmstylethree}%

\def\bbx{{\ensuremath{\mathbf x}}}
\def\bby{{\ensuremath{\mathbf y}}}
\def\bbz{{\ensuremath{\mathbf z}}}
\def\bbW{{\ensuremath{\mathbf W}}}
\def\bbeta{{\mbox{\boldmath $\eta$}}}
\def\bbtheta{{\mbox{\boldmath $\theta$}}}
\def\ccalD{{\ensuremath{\mathcal D}}}
\def\ccalL{{\ensuremath{\mathcal L}}}
\def\ccalR{{\ensuremath{\mathcal R}}}

\begin{document}

\title[Article Title]{Deep Semi-Supervised  Survival Analysis for Predicting Cancer Prognosis }


\author[1]{\fnm{Anchen} \sur{Sun}}\email{asun@miami.edu}
\author[2,3]{\fnm{Zhibin} \sur{Chen}}\email{zchen@med.miami.edu}
\author*[1,3]{\fnm{Xiaodong} \sur{Cai}}\email{x.cai@miami.edu}

\affil[1]{\orgdiv{Department of Electrical and Computer Engineering}, \orgname{University of Miami}, \orgaddress{\street{1251 Memorial Dr.}, \city{Coral Gables}, \postcode{33146}, \state{FL}, \country{USA}}}

\affil[2]{\orgdiv{Department of Microbiology and Immunology}, \orgname{University of Miami}, \orgaddress{\street{1600 NW 10th Ave}, \city{Miami}, \postcode{33136}, \state{FL}, \country{USA}}}

\affil[3]{\orgdiv{Sylvester Comprehensive Cancer Center}, \orgname{University of Miami}, \orgaddress{\street{1600 NW 10th Ave}, \city{Miami}, \postcode{33136}, \state{FL}, \country{USA}}}

\abstract{The Cox Proportional Hazards (PH) model is widely used in survival analysis. Recently,  artificial neural network (ANN)-based Cox-PH models have been developed. However, training these Cox models with high-dimensional features typically requires a substantial number of labeled samples containing information about time-to-event. The limited availability of labeled data for training often constrains the performance of ANN-based Cox models. To address this issue, we employed a deep semi-supervised learning (DSSL) approach to develop single- and multi-modal ANN-based Cox models based on the Mean Teacher (MT) framework, which utilizes both labeled and unlabeled data for training. We applied our model, named Cox-MT, to predict the prognosis of several types of cancer using data from The Cancer Genome Atlas (TCGA). Our single-modal Cox-MT models, utilizing TCGA RNA-seq data or whole slide images, significantly outperformed the existing ANN-based Cox model, Cox-nnet, using the same data set across four types of cancer considered. As the number of unlabeled samples increased, the performance of Cox-MT significantly improved with a given set of labeled data. Furthermore, our multi-modal Cox-MT model demonstrated considerably better performance than the single-modal model. In summary, the Cox-MT model effectively leverages both labeled and unlabeled data to significantly enhance prediction accuracy compared to existing ANN-based Cox models trained solely on labeled data.

 }




\maketitle

\section{Introduction}\label{sec.intro}
Survival analysis deals with time-to-event data, where the event of interest is typically some form of failure or occurrence, such as death, disease relapse, or equipment failure. It has wide applications in areas such as education \cite{singer1993s,ameri2016survival},  engineering \cite{modarres2016reliability},  finance \cite{gepp2008role},  and health care \cite{selvin2008survival,reddy2015review}.   While survival analysis methods are traditionally based on statistical approaches \cite{david2012survival},  machine learning approaches \cite{wang2019machine},  particularly deep learning methods \cite{wiegrebe2024deep},  have also been developed recently.  In medical research, survival analysis methods are often used to analyze the prognosis of diseases \cite{ohno2001modeling}. 

Accurate prediction of cancer prognosis is crucial to the personalized treatment which can be more efficacious, less harmful, and ultimately, more likely to result in successful outcomes than traditional treatment.  Numerous prognostic gene signatures have been developed over last two decades for different types of cancer such as breast cancer \cite{reis2011gene, yu2019breast},  colorectal cancer \cite{lopez2016systematic,ahluwalia2021clinical},  lung cancer \cite{tang2017comprehensive,ge2023systematic}, and prostate cancer \cite{fine2019genomic,li2022comprehensive}.  These signatures use the expression values of several to several hundred selected genes to predict a cancer patient's prognosis.   However,  only a few of them have proven values  in clinical trials, and are in clinical use for a limited set of cancers.     The 21-gene signature Oncotype DX \cite{paik2004multigene} has been widely used in clinical practice for ER-positive breast cancer in the USA.   The 70-gene signature MammaPrint \cite{van2002gene_profiling, van2002gene_signature} has been approved by the US Food and Drug Administration for clinical use.  It provided  good  prognostic values for   ER-positive cancers but classified almost all ER-negative cancers as high risk  \cite{van2002gene_signature}.  These gene signatures typically employ a  simple model to predict cancer prognosis based on expression values of a set of several to several hundreds selected genes, which may have limited their predictive power. 

The Cox proportional hazards model is a popular method in survival analysis to determine the relationship between the survival time of subjects and one or more predictor variables \cite{cox1972regression}.   While the traditional Cox model employs a regression model to predict the hazard ratio,  the Cox model based on an artificial neural network (ANN) have  been  developed  recently  to predict cancer prognosis using the whole transcriptome, bypassing the gene selection step \cite{yousefi2017predicting,katzman2018deepsurv,ching2018cox,qiu2020meta,kim2020improved}.  SurvivalNet \cite{yousefi2017predicting},  DeepSurv \cite{katzman2018deepsurv}, and Cox-nnet \cite{ching2018cox} used a multilayer perceptron (MLP) with one or several hidden layers, and have demonstrated similar performance to the  traditional Cox regression model regularized with the elastic net penalty \cite{simon2011regularization}.  The performance of MLP-based Cox models \cite{yousefi2017predicting,katzman2018deepsurv,ching2018cox} was  limited possibly by the relatively small number of data samples  available to train the model. One way to address this issue is by using a method that can utilize data from multiple types of  cancer.  Towards this end,  a meta-learning method  \cite{qiu2020meta} and a variational autoencoder-based Cox  model (VAECox) \cite{kim2020improved} were developed.  The meta-learning \cite{qiu2020meta} performed similarly  or slightly better  than an MLP  with two hidden layers similar to the MLPs  in \cite{yousefi2017predicting,katzman2018deepsurv,ching2018cox}.    VAECox offered slightly better performance than  Cox-nnet,  although it is unclear if the difference is statistically significant.   
We recently used  deep contrastive learning (CL) \cite{khosla2020supervised,le2020contrastive} to learn features from tumor transcriptomes and then used learned features to train a Cox model to predict cancer prognosis \cite{sun2024deep}. Our CL-based approach offers significantly better prediction accuracy than other existing methods \cite{yousefi2017predicting,katzman2018deepsurv,ching2018cox,qiu2020meta,kim2020improved}. 

All these methods use a supervised learning approach for model training, which requires {\it labeled}  data where each data sample consists of a feature vector containing gene expression values and a target, which is the time to the clinical end point, such as the progression free interval.  
The number of labeled data available for model training is usually small, which may have limited the performance of these ANN-based models.  However, many {\it unlabeled} data that contain tumor gene expression values but with no information about the clinical end point are publicly available in databases such as Gene Expression Omnibus (GEO).  To exploit both labeled and unlabeled data, semi-supervised learning (SSL) methods have been proposed for survival analysis \cite{bair2004semi, shi2011semi, liang2016cancer, chai2017new, shi2021prognostic, roy2022survival, yansari2022new, nateghi2022predicting, chi2021deep, li2023multi},  only two  of which 
 \cite{chi2021deep, li2023multi} are based on ANNs,  but neither the multi-task approach \cite{chi2021deep} nor the multi-view approach   \cite{li2023multi}  is applicable to prediction of cancer prognosis using gene expression values.  

 Recent works have shown that deep semi-supervised learning (DSSL) methods can offer significantly better performance than supervised deep learning with the same amount of labeled data for image classification \cite{tarvainen2017mean,pham2021meta,oliver2018realistic,verma2022interpolation}.  In this paper, we develop a DSSL approach to train the Cox model for survival analysis,  and apply it to predict the prognosis of several types of cancer.  More specifically, we use the DSSL method to  train single-modal Cox models using gene expression data or histological images,  as well as a multi-modal Cox model using both gene expression data and histological images for predicting cancer prognosis.  We demonstrate that our DSSL approach can significantly improve prediction accuracy compared with the existing supervised learning approach.

\section{Results}\label{sec.result}

\subsection{Performance of the single-modal Cox-MT model }
\begin{figure}[!t]
\centering
\includegraphics[width=1\textwidth]{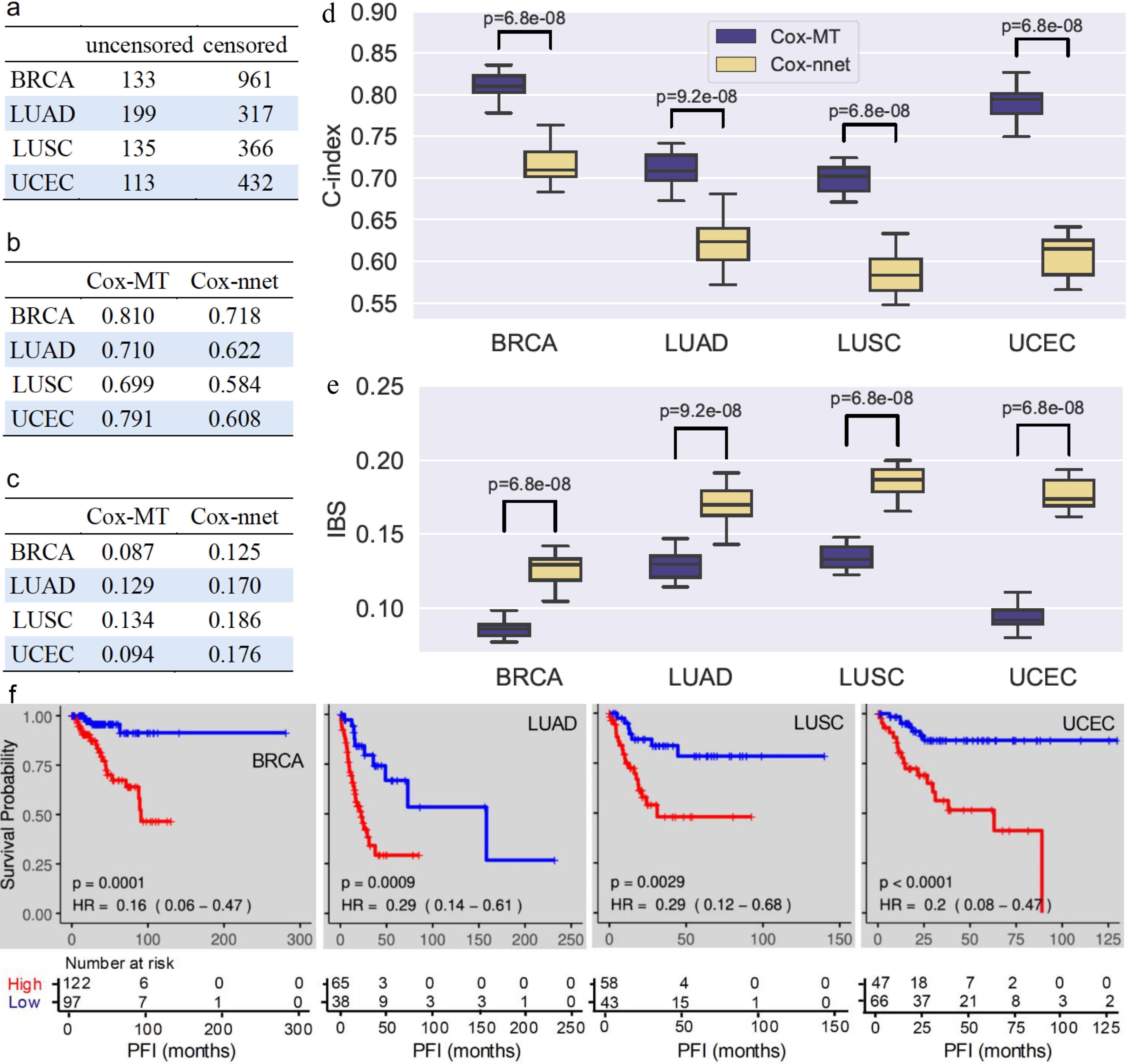}
\caption{\small Performance of single-modal Cox-MT and Cox-nnet models using gene expression values from TCGA to predict the prognosis of four types of cancer.  {\bf a. } Number of samples for each type of cancer. {\bf b.} Average c-indexes of two models. {\bf c. } Average IBSs.  {\bf d.} Box-plots of c-indexes.   The Wilcoxon rank-sum test was used to obtain p-values. {\bf e. } Box-pots of IBSs.   {\bf f.} KM curves for the two groups of patients in the test data stratified by the median HR of the training data predicted by the Cox-MT model: a high risk group (red) characterized by HRs exceeding the median and a low risk group (blue) defined by HRs below the median.   }
\label{fig.cgene}
\end{figure}

Our DSSL-based Cox model,  named as Cox-MT,  was built on the mean teacher (MT) framework originally developed for classification tasks \cite{tarvainen2017mean}.  We trained the Cox-MT model for four types of cancer using the TCGA RNA-seq gene expression data and progression free interval (PFI) data. The four types of cancer are breast invasive carcinoma (BRCA),  lung adenocarcinoma (LUAD),  lung squamous cell carcinoma (LUSC),  and uterine corpus endometrial carcinoma (UCEC).   As shown in Figure \ref{fig.cgene}a, the number of uncensored samples for BRCA, LUAD, LUSC, and UCEC is 133,  199,  135,  and 113,  respectively,  and the number of censored samples is 961, 317, 366  and 432, respectively.  For each type of cancer,  we calculated the variance of expression values of each gene,  and selected 4,000 genes with largest variance as features to train the Cox-MT model.  For comparison, we used the same data to train the Cox-nnet model \cite{ching2018cox}.  

Cross validation was used to determine the values of hyperparameters including the number of layers and the number of nodes at each layer of the MLP in both  Cox-MT and Cox-nnet models.  Cox-MT models for all four types of cancer had two hidden layers.  The number of nodes at the first hidden layer for BRCA, LUAD, LUSC, and UCEC was 1,500,1,000,1,000,1,000,  respectively,  and the number of nodes at the second layer was 200,200,100,2000, respectively.   The Cox-nnet model had one hidden layer with 64 nodes for all four types of cancer.   Figure \ref{fig.cgene}b and d demonstrate the average  Harrell's concordance indexes (c-indexes) of the Cox-MT and Cox-nnet models over twenty runs with random splitting of data into a training set and a test set, and box plots of the c-indexes of these twenty runs.   It is observed that the c-index of Cox-MT is significantly higher than that of Cox-nnet for each type of cancer.  The increase in  the average c-indexes is  0.09,  0.09  0.12,  and 0.18 for BRCA, LUAD, LUSC, and  UCEC, respectively.  Particularly, the c-indexes of Cox-MT for BRCA and UCEC are 0.81 and 0.79, respectively.  Of note, there are three p-values equal to $6.8\times 10^{-8}$ in Figure \ref{fig.cgene}d,  because it is the lowest p-value for the Wilcoxon rank-sum test with 20 samples in the two groups under comparison, where the samples in one group are ranked higher than all samples in the other group as shown by the box plots. 

Figure \ref{fig.cgene}c and e show the average integrated Brier scores (IBSs) and box-plots of IBSs of the Cox-MT and Cox-nnet models over twenty runs.   The IBS of Cox-MT is significantly lower than that of Cox-nnet for each type of cancer.  The decrease in IBS is 0.038, 0.041, 0.052, and 0.082 for BRCA, LUAD, LUSC, and UCEC, respectively.  Figure \ref{fig.cgene}f depicts Kaplan-Meier (KM) curves of the two groups of patients stratified by the median hazard ratio (HR) predicted by the Cox-MT model for four types of cancer. It is seen that the KM curves in the two groups of each type of cancer are significantly different.  

\begin{figure}[!t]
\centering
\includegraphics[width=1\textwidth]{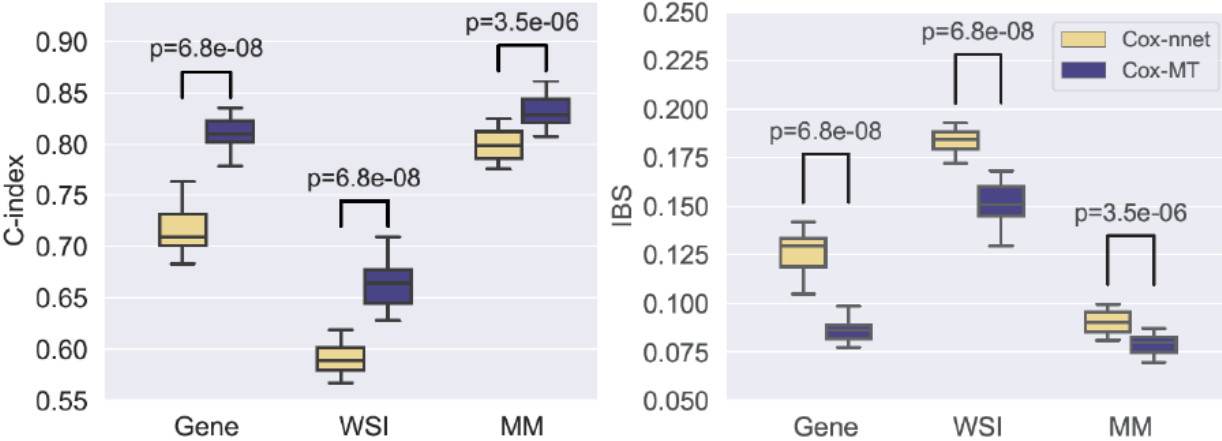}
\caption{{\small Performance comparison for single-modal Cox-MT models using gene expression values or WSIs and  the multi-modal Cox-MT model  using both gene expression values and WSIs of TCGA breast cancer patients.   The x-label MM stands for the multi-modal model.   The Wilcoxon rank-sum test was used to obtain p-values.} }  
\label{fig.mm}
\end{figure}

The performance of single-modal Cox-MT and Cox-nnet models using whole slide images (WSIs) from the TCGA BRCA data set can be found in Figure \ref{fig.mm}.  The BRCA WSI data set contains 124 uncensored samples and 914 censored samples. The Cox-MT model had two hidden layers with 256 nodes at the first layer and 64 nodes at the second layer.  The Cox-nnet model had one hidden layer with 32 nodes.  The average c-index for Cox-MT is 0.66, significantly surpassing the 0.59 observed for Cox-nnet. In a similar vein, the average IBS for Cox-MT is 0.151, which is notably lower than the 0.184 associated with Cox-nnet. Results presented in Figures \ref{fig.cgene} and \ref{fig.mm} clearly demonstrate that Cox-MT significantly outperforms Cox-nnet across both RNA-seq and WSI modalities using the same data set.

\subsection{Impact of the number of unlabeled samples}
\begin{figure}[!t]
\centering
\includegraphics[width=1\textwidth]{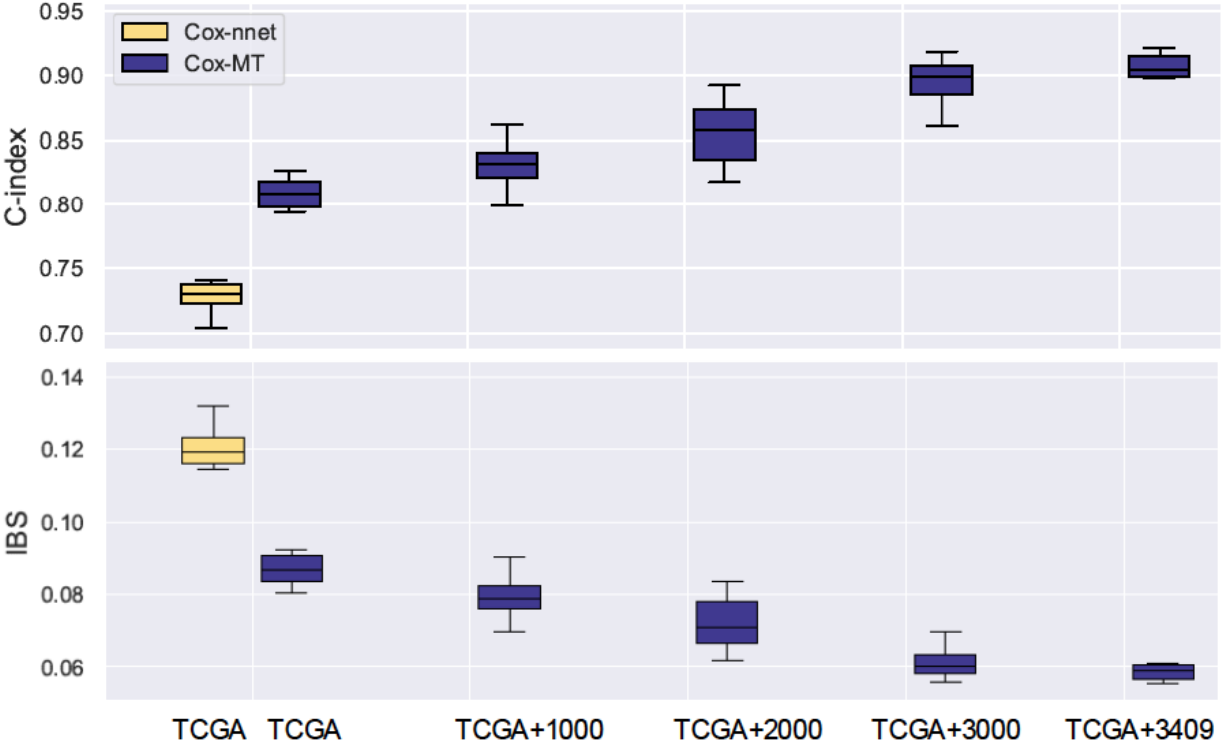}
\caption{\small Performance of the single-modal Cox-MT model  using gene expression values of breast cancer patients with various number of unlabeled data samples.  TCGA+$n$,  $n=1,000, 2,000, 3,000, 3,409$, stands for a data set that includes TCGA BRCA RNA-seq data and additional $n$ unlabeled breast tumors from the GSE96058 data set.  }  
\label{fig.cgene_samples}
\end{figure}

To investigate the number of unlabeled samples on the performance of the Cox-MT model, we used the TCGA BRCA RNA-seq data together with a breast cancer RNA-seq data set in the  Gene Expression Omnibus (GEO) with accession number GSE96058, which contains 3,409 samples.  We formed four data sets by adding 1,000, 2,000, 3,000, and 3,409 samples of the GEO data sets to the TCGA data set as unlabeled samples,  and  then used each data set to train a Cox-MT model.  The performance of the trained models is depicted in Figure \ref{fig.cgene_samples}.  The average c-index of the Cox-MT model for the five data sets,  including the TCGA data and  $n = 0,  1,000, 2,000, 3,000$, and 3,409 extra unlabeled samples,  is 0.81,  0.83,  0.86,  0.89, and 0.90, respectively.  The average IBS of the Cox-MT model for the five data sets is 0.087,   0.079, 0.072, 0.061, and 0.061 respectively.   These results show that increasing the number of unlabeled samples can significantly improve prediction accuracy of the Cox-MT model.

\subsection{Performance of the multi-modal Cox-MT model}
A multi-modal Cox-MT model was trained using TCGA BRCA RNA-seq and WSI data.  As shown in Figure \ref{fig.mtcox}b, our multi-modal Cox-MT model uses a cross-attention mechanism based on the transformer structure to fuse features of two modalities.  To compare the performance of our multi-modal Cox-MT model with that of Cox-nnet, we first projected both the gene expression features and WSI features extracted with DINOv2 to an embedding of 64 dimension,  concatenated the features of two modalities,  and then used the concatenated features to train a Cox-nnet model.  The network structures of Cox-MT and Cox-nnet were determined through cross-validation.  In the Cox-MT model, each multi-head attention unit had four attention heads.  The output of each attention unit had a dimension of 256.  The MLP had  two hidden layers with 1,000 and 200 nodes, respectively.  In the Cox-nnet, the MLP had one hidden layer with 64 nodes. 

The performance of the multi-modal models is presented in Figure \ref{fig.mm} in comparison with the performance of the single-modal model with either RNA-seq data or WSI data.    The multi-modal Cox-MT model achieved the highest average c-index of 0.83, surpassing the multi-modal Cox-nnet at 0.80 (p-value = $3.5\times 10^{-6}$), the single-modal Cox-MT with RNA-seq features at 0.81 (p-value = $4.6\times 10^{-4}$), and the single-modal Cox-MT with WSI features at 0.66 (p-value = $6.8\times 10^{-8}$).  In addition, the multi-modal Cox-MT model achieved the lowest average IBS of 0.079,  outperforming the 0.091 of the multi-modal Cox-nnet (p-value = $3.5\times 10^{-6}$,  the 0.087 of the single-modal Cox-MT with RNA-seq features, (p-value = $4.2\times 10^{-4}$) and  the  0.151 of the single-modal Cox-MT with WSI features(p-value = $6.8\times 10^{-8}$).   

\subsection{Ablation study}
To investigate the influence of various components of Cox-MT on the performance,  we conduct experiments on the TCGA BRCA and GSE96058 RNA-seq data,  varying one  hyperparameter at a time while keeping the others fixed at their default values.  The hyperparameters investigated include the EMA constant $\alpha$,  the consistency weight $w$, student and teacher dropout rates,  and the standard deviation $\sigma$ of Gaussian noise input to the student and teacher models.  The default values of these hyperparameters were $\alpha=0.99$, $w=1$,   $\sigma = 0.1$,  and both student and teacher dropout rates were 0.2. 

\begin{figure}[!t]
\centering
\includegraphics[width=1\textwidth]{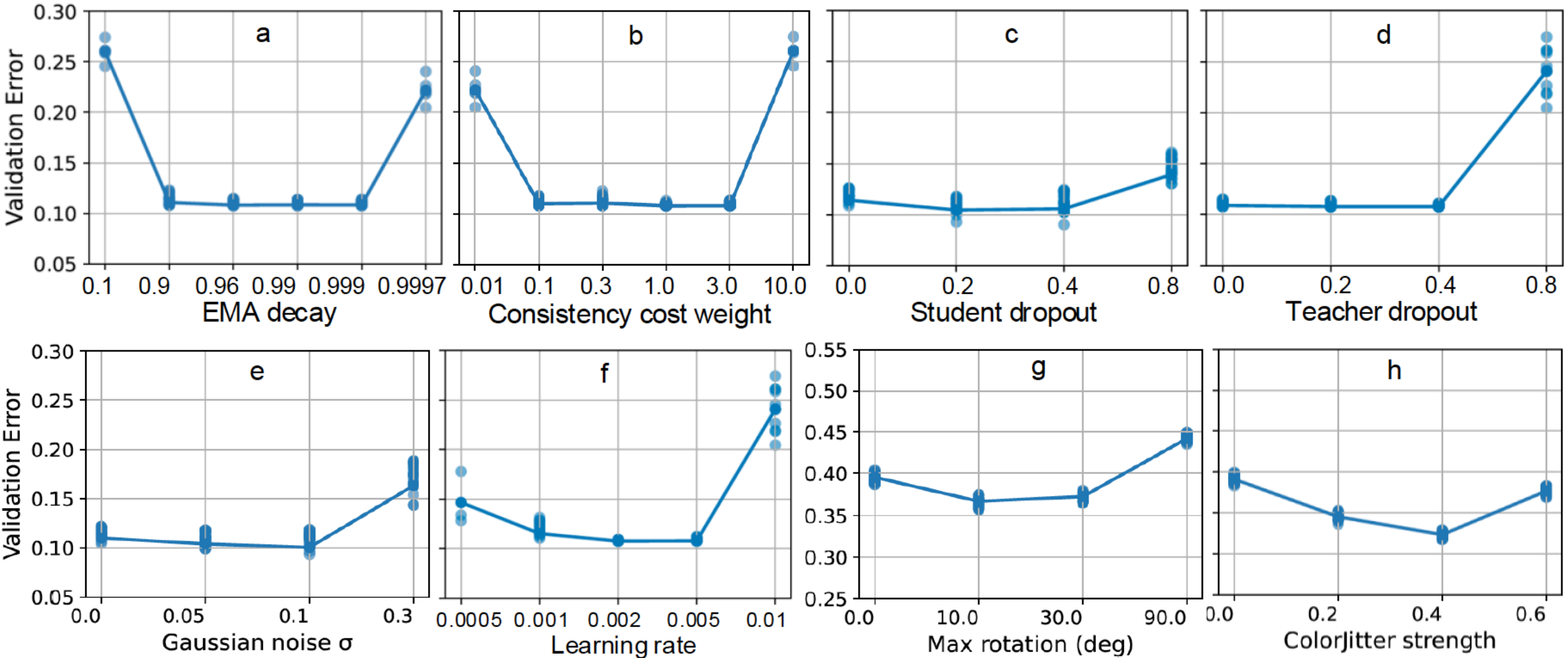}
\caption{{\small Validation errors (1 - c-index) on the TCGA BRCA and GEO GSE96058 RNA-seq data over twenty runs per hyperparameter setting and their means. In each experiment, we varied one hyperparameter, and used default values for the rest.   } }  
\label{fig.ablation}
\end{figure}

 The best values for $\alpha$ are in the range from 0.9 to 0.999 as shown in Figure \ref{fig.ablation}a, and the best values for $w$ are in the interval from 0.1 to 3 as observed from Figure \ref{fig.ablation}b.  Both intervals are relatively large,  implying that the prediction accuracy of Cox-MT is not sensitive to both hyperparameters. 

In the presence of Gaussian noise,  the dropout of both the student and the teacher does not affect the performance much as long as the drop rate is smaller than  or equal to 0.4,  as demonstrated in Figures \ref{fig.ablation}c and d,   However, the performance degrades when the dropout rate is greater than 0.4.  On the other hand,  with the student and teacher dropout rate equal to 0.2,  the optimal values of $\sigma$ ranges from 0 to 0.1, as depicted in \ref{fig.ablation}e.    We also investigated the learning rate on the performance of Cox-MT. As observed from \ref{fig.ablation}f,  a learning rate  in the range from 0.001 to 0.005 achieved the smallest validation error. 

We then carried out experiments on  TCGA BRCA WSI data to investigate the effect of data augmentation on the performance of single-modal Cox-MT model using WSI data.    We considered two types of data augmentation for imaging data: random rotation and color jitter, and applied either of them, but not both, to the patches of WSIs.  For random rotation, we varied the maximum degree of the rotation $\beta$.  For color jitter, we set the brightness, the contrast, and the saturation to the same value $\gamma$ and hue to 0.1, and then varied the value of $\gamma$.   The other hyperparameters, $\alpha$, $w$,  and student and teacher dropout rates, were set to their default values; $\sigma$ was set to zero.  
As shown in Figure \ref{fig.ablation}g,  the range of optimal values for $\beta$ is  $10^o$ to $30^o$,  and as observed in  Figure \ref{fig.ablation}h,  the optimal values of $\gamma$ is around 0.4.  Interestingly,  color jitter provided better performance than random rotation.

\section{Discussion}\label{sec.dis}
While ANNs can learn effective feature representation from complex data for specific prediction tasks,   training  an ANN  with high-dimensional features typically require a large number of labeled data samples.  In the case of supervised training of ANN-based Cox  models,  the labeled data include the information about the features and the time-to-event.  However,  collecting the time-to-event information in the clinical setting from, e.g. cancer patients, is challenging, because it requires long follow-up time. Therefore, the labeled data for survival analysis in clinical applications are usually scarce.   This constraint  limits the prediction accuracy of the trained model.   On the other hand,  semi-supervised learning can leverage both labeled and unlabeled data to train a model,  and the use of additional unlabeled data can improve the performance of the trained models.  

In survival analysis, our DSSL-based Cox-MT model significantly outperforms Cox-nnet using the same data set that includes both uncensored and censored samples without  unlabeled samples that are not associated with any time-to-event information.  This is due to the fact that Cox-MT can exploit the information in censored samples better than Cox-nnet.  In training Cox-nnet,  the negative partial likelihood function of the data $\ccalL_s$,  that is the negative likelihood of the uncensored samples, is used as the loss function.  In contrast,  the loss function in training Cox-MT is $\ccalL= \ccalL_s + w \ccalL_u$, where $\ccalL_u$ is based on censored and unlabeled samples.  Therefore, even without unlabeled samples,  censored samples can help Cox-MT improve performance.  When unlabeled samples are included in training Cox-MT,  our results show that the performance of Cox-MT can be further improved.  In fact, the performance improvement increases as the number of unlabeled samples increases, given the same set of uncensored and censored data.  While our single-modal Cox-MT outperforms Cox-nnet across two modalities of gene expression value and WSI,   our multi-modal Cox-MT also offer better performance than the multi-modal Cox-nnet and the single-modal Cox-MT.   

While we applied our Cox-MT model to the  cancer survival data, our model is apparently applicable to the survival analysis in the other fields such as education and  finance.  Our DSSL-based Cox-MT model is expected to provide significantly better performance than the traditional supervised learning-based Cox model in various applications.

\section{Methods}\label{sec.methods}

\subsection{Data sets}
We used RNA-seq gene expression and clinical data of four types of cancer, BRCA, LUAD, LUSC, and UCEC, from The Cancer Genome Atlas (TCGA).   RNA-seq data were downloaded from the data portal of Genomic Data Commons (GDC) in the National Cancer Institute,  and the  expression values in term of Fragments Per Kilobase of transcript per Million mapped reads upper quartile (FPKM-UQ)  were extracted from the RNA-seq data.  The corresponding clinical data of the cancer patients curated by the PanCancer Atlas consortium \cite{liu2018integrated} were downloaded from the GDC Pan-Caner Atlas website (https://gdc.cancer.gov/about-data/publications/pancanatlas).  The PFI and the censoring information were extracted from the clinical data.  Of note, we used PFI instead of the overall survival (OS) time in our
analysis, because PFI is generally a better clinical endpoint choice than OS \cite{liu2018integrated}. Genes with missing expression values were discarded.  For each type of cancer, we calculated the variance of the expression values of each gene, and expression values of the top 4,000 genes with the largest variance were extracted and used to train and test Cox-MT models.  Let the expression value of a gene be denoted as $x$, then, it was log-transformed as $\log_2(1 + x)$ before it was used in model training and testing.

Another set of  RNA-seq gene expression data of breast cancer patients was downloaded from the Gene Expression Omnibus (GEO)
with accession number GSE96058  \cite{brueffer2018clinical}.  It contains 3,273 samples and 136 technical replicates,  and was used as unlabeled data together with the TCGA BRCA data to train COX-MT models.  The FPKM values in this data set were normalized to the TCGA BRCA data as follows.   Based on RNA-seq data,  Eisenberg and Levanon identified eleven house keeping genes that  were highly uniform and strongly expressed in all human tissues  \cite{eisenberg2013human}.  Ten out  of the eleven genes are present in the TCGA data set, and they are  C1orf43,  CHMP2A,   GPI,  PSMB2,  PSMB4, RAB7A,  REEP5,  SNRPD3,  VCP,   and VPS29.     We computed the average expression values of these 10 genes in the TCGA BRCA and GSE96058 data sets,  which are denoted as $E_t$ and $E_g$, respectively.    Then, the expression values of all genes in the GSE96058 data set were multiplied by a normalization factor $E_t/E_g$. 

The WSI data for TCGA BRCA patients were sourced from The Cancer Imaging Archive. This data, along with the TCGA BRCA clinical information, was utilized to train a single-modal Cox-MT model, as well as a multi-modal Cox-MT model that incorporated the TCGA BRCA RNA-seq data.

\subsection{Survival model}  In survival analysis, the hazard function $h(t)$  represents the instantaneous rate at which events occur for individuals who  have survived up to time $t$.  Survival models capture a relationship between the duration of time until an event occurs and one or more covariates that may influence that time.  A commonly used survival model is the Cox-PH model \cite{cox1972regression}, which specifies $h(t)$  as follows $h(t,\bbx) = h_0(t)\exp(\bbtheta^T\bbx)$,  where the $d\times 1$ vector $\bbx$ contains $d$ covariates or features that may influence the event time,  vector  $\bbtheta$ represents the model parameters,  $h_0(t)$ is the baseline hazard, and we have explicitly written $h(t)$ as $h(t; \bbx)$, a function of both $t$ and the feature vector $\bbx$.  In this paper, we use an  ANN, denoted as $f_{\theta}(\bbx)$, instead of the original linear model to estimate the hazard ratio, and thus the Cox-PH model can be represented as  $h(t,\bbx) = h_0(t)\exp(f_{\theta}(\bbx))$.

\subsection{Single-modal Cox-MT model}
Consider a data set available for survival analysis,  where each sample is characterized by a $d\times 1$ feature vector $\bbx_i$ representing the values of $d$ features or variables that may  influence the time-to-event for subject $i$.   The data set includes two subsets,  one uncensored event subset $\ccalD_e$  where each sample has a time-to-event $t_i$,   and the other censored subset $\ccalD_c$ where each sample  has a time  $t_i$ to a point when no event occurs.  It may include  another unlabeled subset $\ccalD_u$ where each sample does not have any time information.  In the case of using gene expression values to predict cancer prognosis, the feature vector $\bbx_i$ contains gene expression values  of  a tumor from the $i$th patient, and $t_i$ can be the OS time or PFI. 

\begin{figure}[!t]
    \centering
    \includegraphics[width = 0.8\linewidth]{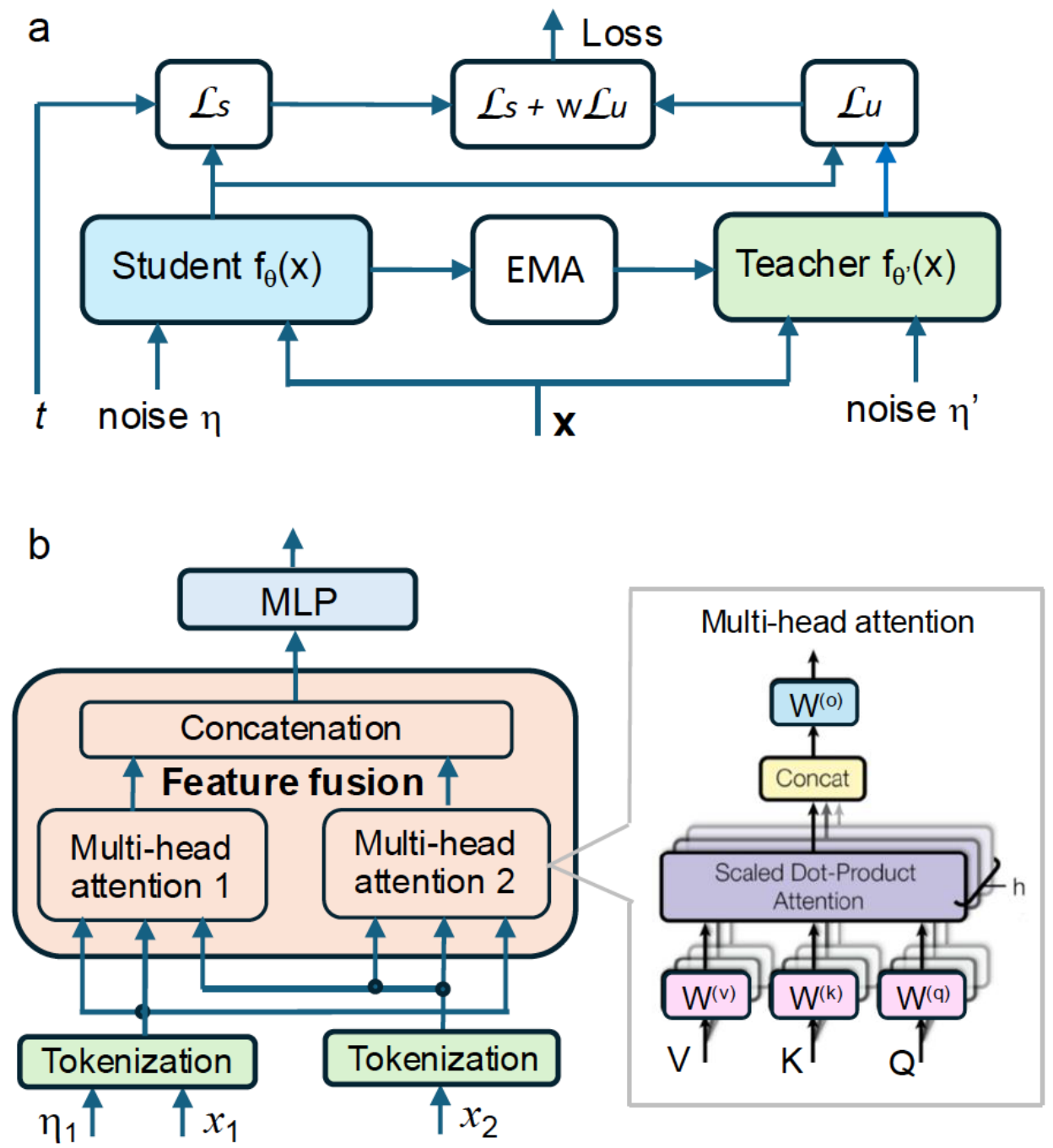}
\caption{{\bf Cox-MT models for predicting cancer prognosis.} {\bf a.} Single-model Cox-MT model.  {\bf b.} The student model of the multi-modal Cox-MT model which has the same overall structure as the single-modal Cox-MT model.  }
\label{fig.mtcox}
\end{figure}

We will use the framework of the mean teacher model  \cite{tarvainen2017mean},  that is a DSSL method for image classifications,  for  our  DSSL-based Cox model that is named as the Cox-MT model.  It is  depicted in  Figure \ref{fig.mtcox}a, where $f_\theta(\bbx)$, named as a student,  stands for an MLP with $\theta$ representing all weights and bias,    and $f_{\theta'}(\bbx)$, named as a teacher,  is another MLP with the same structure as the student but with different weights and bias $\theta'$. The EMA block stands for  exponential moving average, which computes the teacher's weights at the $t$th step of training as $\theta'_t= \alpha \theta'_{t-1}+(1-\alpha)\theta_t$, where $0< \alpha <1$.  Both MLPs have one node at the output layer.  

Of note,  the input data $\bbx$ in  Figure \ref{fig.mtcox}a represents both labeled and unlabeled samples, while $t$ represents the time to event of the uncensored samples or the time to censoring of the censored samples. We define a supervised learning loss function $\ccalL_s$ as the negative log partial likelihood of the uncensored data:
\begin{equation}
 \ccalL_s= -\frac{1}{|\ccalD_e|}\sum_{\bbx_i\in\ccalD_e} \Bigl[ f_\theta(\bbx_i+\bbeta_i) -\log\bigl(\sum_{\bbx_j\in \ccalR(t_i)} \exp(f_\theta(\bbx_j+\bbeta_j)\bigr)    \Bigr],
\end{equation}
 where $|\ccalD_e|$  represents the size of $\ccalD_e$,   $\bbeta_i$ and $\bbeta_j$ are zero mean Gaussian noise,  and $\ccalR(t_i)\subset \ccalD_e\cup \ccalD_c$ is the set
of individuals that are at risk of experiencing an event at time $t_i$.  We also define a loss function $\ccalL_u$ for the unlabeled samples and censored samples as follows:
\begin{equation}
\ccalL_u=\frac{1}{|\ccalD_u\cup\ccalD_c|}\sum_{\bbx\in \ccalD_u\cup \ccalD_c}\Bigl[f_\theta(\bbx+\bbeta)-f_{\theta'}(\bbx+\bbeta')\Bigr]^2,
\end{equation}
where   $|\ccalD_u\cup\ccalD_c|$ represent the size of the union of $\ccalD_u$ and $\ccalD_c$,   $\bbeta$ and $\bbeta'$ are zero mean Gaussian noise.  

We trained the model with the following loss: $\ccalL = \ccalL_s + w \ccalL_u$,  where $w$ is a positive constant that balances the two losses.  If the data size is relatively small, we can use all the data samples to form $\ccalL_s$ and $\ccalL_u$ and train the model.  If the data size is relatively large,  and more efficient stochastic gradient descent (SGD) algorithm  or its variant is needed to  train the model,   we need to carefully sample data for $R(t_i)$ to  form mini-batches \cite{kvamme2019time}. 

We investigated the single-modal Cox-MT model with two types of input data for predicting cancer prognosis:  one used  gene expression data and the other used whole slide images (WSIs).   As mentioned earlier, the gene expression data can be represented as a feature vector $\bbx$ and input to the model as shown in Figure \ref{fig.mtcox}a.   For WSI data,  we extracted features from the images and then input the features to Cox-MT.  Specifically, the  WSI of a tumor from a breast patient was processed with U-Net to extract the tumor region,  which is then divided into patches of $224 \times 224$ pixels. Each patch of the same WSI was input to  a DINOv2 model \cite{oquab2023dinov2} pretrained with ImageNet data, and the feature vectors output by the DINOv2 model for all patches of the same WSI were averaged to form a feature vector $\bbx$ which was input to the student model.    An augmented version of each patch was generated,  and input to DINOv2; features vectors  output by DINOv2 for all augmented patches of the same WSI were again averaged to form a feature vector $\bbx'$ which was input to the teacher model. 

 \subsection{Multi-modal Cox-MT model}
 We developed a multi-modal Cox-MT model that accepted both gene expression and WSI data as input.  The multi-modal model has the same overall structure as the single-modal model in Figure \ref{fig.mtcox}a, but with different student and teacher models.  The student model of the multi-modal Cox-MT model is depicted in Figure \ref{fig.mtcox}b.  The teacher model is again obtained from the student model with EMA.  Let $\bbx_1$ be the feature vector that contains gene expressions,  and $\bbx_2$  contains the features extracted from a WSI using DINOv2 as described earlier.    Two features $\bbx_1$  and $\bbx_2$   are fused with a mutual attention block that includes two multi-head attention (MHA) units.   
 
 First, $\bbx_1$  and $\bbx_2$   are tokenized as follows.  Gaussian noise $\bbeta_1$ was added to $\bbx_1$, resulting in $\bby_1=\bbx_1+  \bbeta_1$.  The tokenization block generates a sequence of 32 tokens as follows $\bbz_1^{(j)}=\bby_1\bbW_1^{(j)}$,  $ j=1,2,\cdots,32$,  where $\bbW_1^{(j)}$ is a $4,000\times 256$  linear projection matrix that is learnable.  As mentioned earlier,  the WSI of a patient was divided into $n$ patches of $224 \times 224$ pixels,  $\bbx_2^{(j)}, j =1,2,\cdots, n$, which were input to  the pretrained DINOv2 model.  Taking the $1\times 1,024$  CLS token from the output of DINOv2 as the extracted features from each patch, we obtained a sequence of $n$ feature vectors $\bby_2^{(j)}$, $j=1,2,\cdots, n$.   A sequence of tokens was obtained as follows: $\bbz_2^{(j)} = \bby_2^{(j)} \times \bbW_2$, where $\bbW_2$ is a $1,024\times 256$ projection matrix. If $n$ is greater than or equal to 128,  the tokenization unit output 128 randomly selected tokens. If $n$ is less than 128, then zero tokens were padded so that the number of tokens output by tokenization unit  was 128.    A CLS token was prepended to each token sequence. The two token sequences are input to the two MHA units, and the queries of two MHA units are swapped, which is referred to as mutual attention  \cite{cai2023multimodal}.  In other words,  the query for MHA$_1$ is generated from the second token sequence, and similarly the query for MHA$_2$ is generated from the first token sequence.   The outputs of two MHA units are concatenated and input to an MLP.  Of note,   $\bbx_1$ for samples in $\ccalD_c\cup\ccalD_u$ is input to the first attention block of the teacher model along with a Gaussian noise $ \bbeta_1'$,  and $\tilde \bbx_2^{(j)}$,  that is generated from augmented version of WSI patches generating $\bbx_2^{(j)}$, is input to the second attention block of the teacher model.

 \subsection{Model training and performance evaluation}
The data set $\ccalD_e\cup\ccalD_c$ was randomly divided into five equal folds. One fold was designated as the test set, and the remaining four folds were designated as the training and validation set, which was further divided into a training set with 80\% samples of the  training and validation set and a validation set with 20\% samples.    If an unlabeled set $\ccalD_u$ is available, it  was also randomly divided into five equal folds, and four of which were added to the training set.  The training set was used to trained a Cox-MT model and the performance of the model was evaluated with the validation set.  This training and validation process was carried out over a set of values of hyperparameters that include the learning rate,  the dropout rate, the EMA constant $\alpha$,  the noise variance, and  the network structure.  The optimal values of hyperparameters were determined based on the model performance on the validation set,  and were used to train the Cox-MT model with all samples in the  training and validation data set.  The test data was applied to the trained model to assess the performance.   Model training and testing were carried out five times so that all data samples were used as the test data, although the test data were never used in training.  Moreover, the aforementioned process of randomly splitting data into a test set and a  training and validation set, and training and testing the model was repeated four times.  Therefore, we obtained twenty results for each performance measure, and their average and box plots are reported.

The performance of Cox models on prognosis prediction was evaluated with two commonly used metrics:  the  c-index  \cite{harrell1996multivariable} and the IBS \cite{graf1999assessment}. The c-index is defined as the ratio of the concordant pairs of predictions and all comparable pairs of patients, where a pair of patients is called concordant if the risk of the event predicted by a model is lower for the patient who experiences the event at a later time point. A c-index value of 0.5 indicates  random prediction. The value of c-index increases when the prediction accuracy increases, and a value of 1.0 indicates perfect prediction, where all pairs are concordant.  

Let $T$ be the time to event,  then the survival function $S(t)=P(T>t)$.  The Brier score at time $t$ under random censorship can be estimated as~\cite{graf1999assessment}: 
$BS(t)=\frac{1}{|\ccalD_e\cup\ccalD_c|}\Bigl[   \sum_{i\in\ccalD_e} {\hat S_i^2(t) 1_{t_i\le t }}/{\hat G(t_i)}
					+\sum_{i\in \ccalD_e\cup\ccalD_c} {(1-\hat S_i(t))^21_{t_i>t}}/{\hat G(t)}  \Bigr]$
where $\hat S_i(t)$ is the estimated survival function of individual $i$,  $\hat G(t)$ is the Kaplan-Meier estimated of the censoring distribution,  $1_{t_i>t}=1$ if $t_i>t$ or =0 if  $t_i\le t$.  The survival function   $S_i(t)$ can be estimated as  $\hat S_i(t)= \exp[ - \hat H_0(t) \exp(f_\theta(\bbx_i)) ]$ , where $\hat H_0(t)$ is the cumulative baseline hazard function estimated from the training data using the Breslow estimator \cite{breslow1975analysis}: $\hat H_0(t)=\sum_{j\in\ccalD_e,t_j\le t}  {1}/{\sum_{k\in\ccalR(t_j)}\exp(f_\theta(\bbx_k))}  $. The IBS is then calculated as ${\rm IBS}=\frac{1}{T}\int_0^{T} BS(t) dt$,  where $T$ is determined as follows.  It is seen from the formula of $\hat H_0(t)$ that we can get $\hat H_0(t)$ up to the maximum uncensored $t_j$ in the training data.  However,  the size of the set $\ccalR(t_j)$ decreases when $t_j$ increases.  For those $t$ where the formula of $\hat H_0(t)$ contains $\ccalR(t_j)$  whose size is small,  $\hat H_0(t)$ may not be accurate.  To avoid this problem, we define $t_{\max}$ as the maximum uncensored $t_j$ in the training data where  the size of  $\ccalR(t_j)$ is greater than or equal to 20.  Also, we define $\tilde t_{max}$ as the maximum $t_j$ in the test data.  Then,  we set $T=\min\{t_{\max}, \tilde t_{\max} \}$.

\section*{Data availability }
All data sets used in this paper are publicly available. The TCGA RNA-seq data can be downloaded from the GDC data portal at https://portal.gdc.cancer.gov/,  and the corresponding clinical data of the cancer patients are accessible  from the GDC Pan-Caner Atlas website (https://gdc.cancer.gov/about-data/publications/pancanatlas).  The TCGA breast cancer WSI data can be downloaded from The Cancer  Imaging Archive at https://www.cancerimagingarchive.net/.   Another breast cancer RNA-seq data set is accessible from GEO with an accession number GSE96058.

\section*{Code availability} 
The computer code for reproducing this work is available via GitHub at https://github.com/CaixdLab/CoxMT.





\backmatter

\bmhead{Author contributions}
X.C. conceptualized and designed the study. A.S. developed computer programs and trained the models. Z.C, A.S, and X.C interpreted the data. X.C. drafted the manuscript; all authors reviewed and approved the manuscript.

\bmhead{Funding}
Research reported in this publication was supported by the Sylvester Comprehensive Cancer Center
in partnership with the University of Miami College of Engineering under the Engineering Cancer
Cures program Award Number SCCC-ECC-2022-01.

\bmhead{Competing interests}
All authors declare no competing interests.

\newpage

\newpage

\newpage

\newpage

\newpage

\end{document}